\documentclass[conference]{IEEEtran}
\IEEEoverridecommandlockouts

\usepackage{caption}
\usepackage{subcaption}
\usepackage{subcaption} 
\usepackage{amsmath,graphicx}
\usepackage{cite}
\usepackage{amsmath,amssymb,amsfonts}
\usepackage{algorithmic}
\usepackage{graphicx}
\usepackage[dvipsnames]{xcolor}
\usepackage{amssymb}
\usepackage{textcomp}
\usepackage{xcolor}
\usepackage{blindtext}
\usepackage{multirow}
\usepackage{array}
\usepackage{multicol}
\usepackage{times}
\usepackage[dvipsnames]{xcolor}
\usepackage{amsmath}
\usepackage{amssymb}
\usepackage{booktabs}
\usepackage{caption}
\usepackage{dirtytalk}
\usepackage{balance}
\usepackage{multirow}
\usepackage{color}
\usepackage{cite}

\def\BibTeX{{\rm B\kern-.05em{\sc i\kern-.025em b}\kern-.08em
    T\kern-.1667em\lower.7ex\hbox{E}\kern-.125emX}}

\title{Self-Supervised Learning-Based Multimodal Prediction on Prosocial Behavior Intentions}

\author{%
  \IEEEauthorblockN{%
    \parbox{\linewidth}{\centering
      Abinay Reddy Naini\IEEEauthorrefmark{1}\IEEEauthorrefmark{3},
      Zhaobo K. Zheng\IEEEauthorrefmark{2},
      Teruhisa Misu\IEEEauthorrefmark{2},
      Kumar Akash\IEEEauthorrefmark{2}
    }%
  }%
  \IEEEauthorblockA{%
    \IEEEauthorrefmark{1}University of Texas at Dallas, Dallas, Texas, USA\\
    \IEEEauthorrefmark{2}Honda Research Institute USA, Inc., San Jose, California, USA\\
    \IEEEauthorrefmark{3}Work completed during internship at Honda Research Institute USA%
  }%
}

\begin{document}

\maketitle

\begin{abstract}
Human state detection and behavior prediction have seen significant advancements with the rise of machine learning and multimodal sensing technologies. However, predicting prosocial behavior intentions in mobility scenarios, such as helping others on the road, is an underexplored area. Current research faces a major limitation—there are no large, labeled datasets available for prosocial behavior, and small-scale datasets make it difficult to train deep-learning models effectively. To overcome this, we propose a self-supervised learning approach that harnesses multi-modal data from existing physiological and behavioral datasets. By pre-training our model on diverse tasks and fine-tuning it with a smaller, manually labeled prosocial behavior dataset, we significantly enhance its performance. This method addresses the data scarcity issue, providing a more effective benchmark for prosocial behavior prediction, and offering valuable insights for improving intelligent vehicle systems and human-machine interaction.
\end{abstract}


\section{Introduction}
\label{sec:intro}
Advances in machine learning, wearable sensing, and multimodal fusion enabled and empowered human state sensing and behavior predictions. Such innovations have applications in intelligent vehicles \cite{zepf2020driver}, digital health \cite{garcia2018mental}, and emotion recognition \cite{abdullah2021multimodal}. User behavior prediction is critical for safe and smooth human-machine interaction \cite{jupalle2022automation}, especially for interactions in mobility. Popular applications include takeover prediction in automated vehicles (AV) \cite{pakdamanian2021deeptake}, driving style preference of AV \cite{koochaki2023learn}, and driver emotion prediction \cite{bethge2021vemotion}. This research area is becoming more prominent, with many promising results.

Despite the potential, most ongoing research is still heavily limited by training sample size. Multimodal data collection with human state labels is expensive, and most research only utilizes a small group of participants, ranging from 12 to 50 \cite{morales2020automated}, resulting in less than 2000 training samples. These small datasets prevent the successful training of deep learning models such as Transformers and CNN. Existing literature has extensively demonstrated this limitation, that basic models like Random Forest have outperformed larger-parameter deep neural networks \cite{zheng2022detection}. Thus, appropriate benchmarks for human state detection are missing, especially for novel human states where no public datasets are available. 

It is an emerging field to investigate prosocial behaviors, and people's motivation and tendency to conduct prosocial behaviors in mobility \cite{chi2024should}. Prosocial behaviors intend to benefit other people or society, such as yielding and slowing down for vulnerable road users \cite{ward2020traffic}. Users in modern society care more about well-being and harmony in on-road interactions because the rapid progress in AV and advanced driving assistance systems (ADAS) are fulfilling their needs for basic safety and comfort. Prosocial behavior intention prediction will be fundamentally helpful to encourage prosocial behaviors in transportation and design automated vehicle features to help other road users according to user preference. However, there is not yet a data-driven method to predict such intention.

\begin{figure}[t]
    \centering
    \begin{subfigure}[]{0.14\textwidth}
        \includegraphics[width=\linewidth]{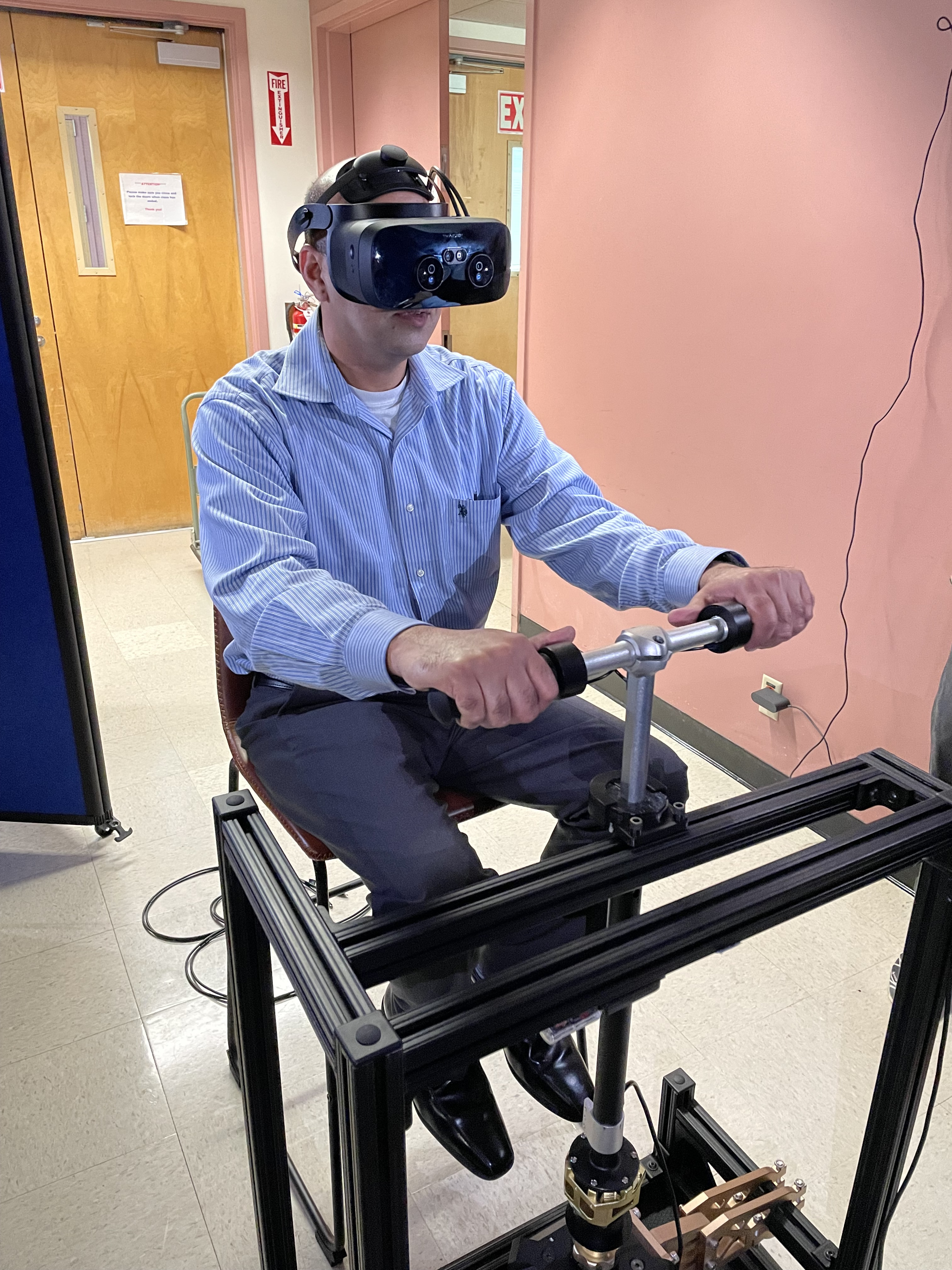}
        \caption{Hardware Setup}
        \label{simulatorsetup}
    \end{subfigure}
    \begin{subfigure}[]{0.33\textwidth}
        \includegraphics[width=\linewidth]{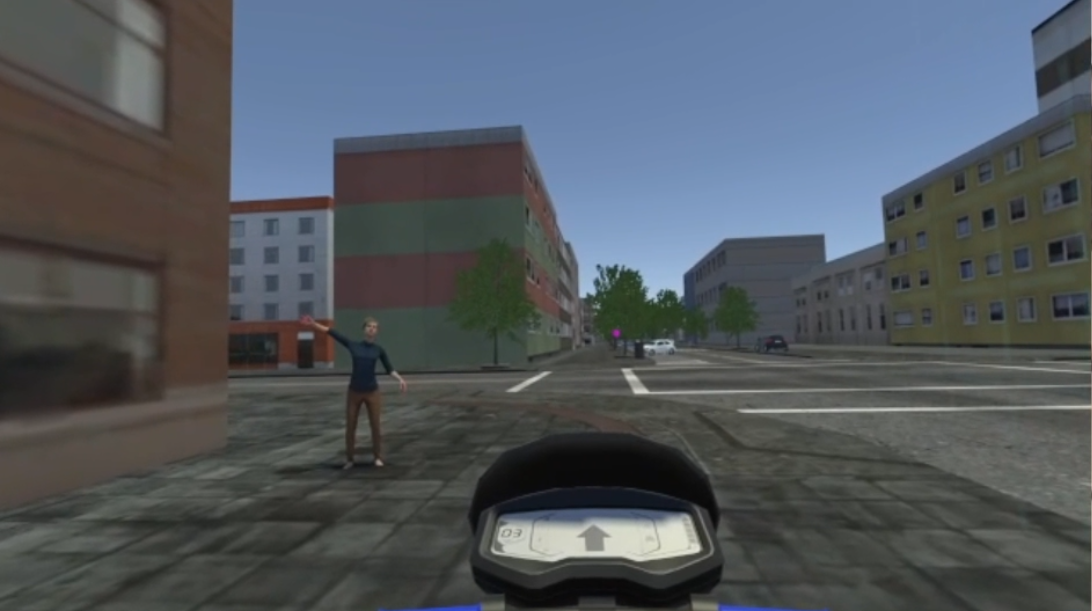}
        \caption{Example VR View}
        \label{vrview}
    \end{subfigure}
    \caption{VR Simulator used for user study}
    \vspace{-0.8cm}
\end{figure}


Building on the challenges of limited labeled datasets for prosocial behavior prediction, it is important to note that while there is a shortage of long, labeled datasets specifically for prosocial behavior prediction, a wealth of datasets focus on different tasks, particularly utilizing physiological data and sensor inputs. Physiological data has been proven effective in detecting human behaviors and cognitive states \cite{andreassi2010psychophysiology}. Though not directly targeting prosocial behavior, these datasets can be invaluable in a self-supervised learning (SSL) approach, where models can pre-train on large, unlabeled, or differently labeled datasets and later fine-tune them for the specific task of prosocial behavior prediction. Such approaches have shown significant promise in related fields, enabling models to generalize well even in low-data scenarios \cite{Chen_2020_2, Chen_2022, Baevski_2020, Naini_2024, Wagner_2023, Goncalves_2024}. Thus, we propose a novel self-supervised pre-trained model that leverages multi-modal datasets from various domains, including physiological data like heart rate, galvanic skin response (GSR), and pupil dilation. By pre-training on diverse datasets from tasks related to emotion recognition, human state sensing, and behavioral prediction, we create a robust foundation that can then be fine-tuned with smaller, manually labeled datasets specific to prosocial behavior intention. This SSL-based approach addresses the current data scarcity challenge and provides a benchmark methodology to enhance model performance across small, domain-specific datasets. The rest of the paper is organized as follows: Section II introduces the dataset and the user study; Section III introduces the machine learning pipeline; Section IV presents the machine learning results on prosocial behavior intention prediction and also its impact on other existing behavioral predictions. We finally discuss the work and conclude the paper in Section V.  

\section{Datasets}
\label{sec:format}
Since there are no public datasets on prosocial behaviors, we designed a user study to collect data. The study intends to present prosocial behavioral chances in an urban traffic environment to participants, and they have the choice to do them or not. We designed a Virtual Reality (VR) simulator, with scooter hardware, as shown in Fig. \ref{simulatorsetup}. We chose the scooter as it allows more social interactions with other road users. The participant wears a Varjo XR-3 VR headset with a wide 115-degree field of view. We used Unity to render the city environment, with other pedestrians, static cars, delivery robots, buildings, and trees in their sight. An example VR scene is shown in Fig. \ref{vrview}, where another pedestrian is waving to the participant for help with navigation. The participants manually drove through the city, and 8 prosocial chances were present. These chances include yielding to other road users, others asking for navigation, slowing down in a busy street, taking a detour to avoid busy streets, and blocking the way of kids running away from their parents. A tutorial session before the experimental sessions lets the participants get familiar with the system such that they interact naturally with the VR road users. An experimenter, not in sight of the participant, continuously observes the participant through a camera. Whenever a prosocial behavior is conducted, the experimenter triggers a corresponding pre-programmed feedback response in the VR simulation for the prosocial act. For example, in Fig. \ref{vrview}, the lady will thank the participant if given navigation information.                

During the user study, we collect multi-modal signals including behavioral and physiological measures from the participants. Gaze and pupil data is collected though the integrated eye tracker in the VR headset. Using ray casting in Unity, we record the object type in their gaze direction, for example, pedestrians or cars. We collect the peripheral physiological signals with an Empatica E4 wristband for heart rate (HR) and a Shimmer GSR+ for galvanic skin response. E4 also has GSR measurement, but we use the Shimmer to lower motion deficits. We utilize a Kinect V2 placed in front of the participant to detect body 3D joint positions. The study was approved by the San Jose State University IRB. 50 participants enrolled in our study. Participants conducted 317 prosocial behaviors out of a total 538 prosocial chances, which is a balance ratio of 59:41. 

We also utilize some publicly available datasets with similar sensing modalities, for our self-supervised model training. The datasets include: 1) takeover prediction dataset in cars and micro-mobilities, with GSR and gaze measurements \cite{zheng2024dualtake}; 2) pilot workload estimation dataset in flight simulation, with GSR, HR, and gaze measurements \cite{park2024pilot}; 3) trust prediction dataset using attention network with GSR and gaze measurements \cite{niu2024beyond}; and 4) driving style preference dataset in automated vehicles, with GSR, HR, and gaze measurements \cite{koochaki2023learn}. We utilize these datasets because they have similar sensing modalities, and thus need no data adaption. These public datasets were originally collected for various other cognitive machine learning tasks; we will use them for the self-supervised learning approach.

The feature set used in the model includes both physiological and body movement features, all Z-normalized with a moving window of 5 minutes. The physiological features include heart rate (1-dim), GSR (1-dim), and pupil diameter (1-dim). The body movement features include Head position (3-dim), Head rotation (4-dim), ShoulderLeft (3-dim), ShoulderRight (3-dim), ElbowLeft (3-dim), ElbowRight (3-dim), WristLeft (3-dim), WristRight (3-dim), HandLeft (3-dim), and HandRight (3-dim). These features capture essential cues for prosocial behavior prediction.

\section{Model Pipeline}
\label{sec:typestyle}

\begin{figure}[t]
	\begin{center}	
		\centerline{\includegraphics[width=0.95\columnwidth]{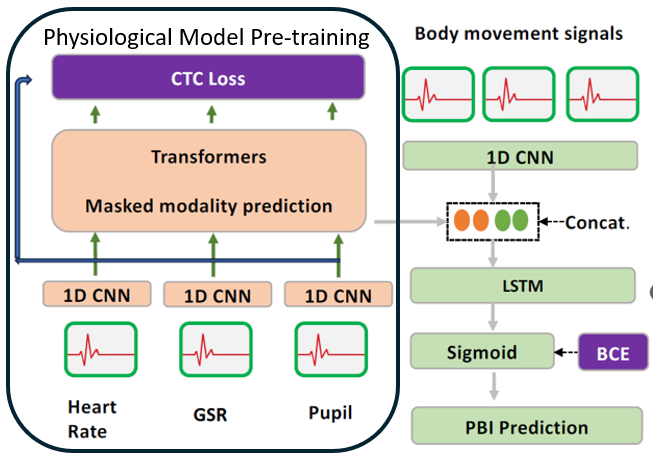}} 
		\vspace{-0.2cm}
		\caption{Block diagram of the building blocks of our proposed approach to predict prosocial behavior intention (PBI).}
		\label{fig:block1}
	\end{center}
	\vspace{-1cm}
\end{figure}

Figure \ref{fig:block1} shows our proposed framework consisting of SSL based pre-trained model followed by a prediction of prosocial behavior intention training. This section describes each of the proposed model pipelines in detail. 

\subsection{Pre-Trained physiological model}
\label{sec:pagestyle}

Figure \ref{fig:block1} illustrates the architecture of our pre-trained physiological model, which is the first step in the proposed framework. The model takes three input physiological modalities: heart rate, GSR, and pupil diameter. Each of these modalities is passed through a dedicated 1D CNN to extract low-level feature representations, ensuring that the model can process each physiological signal independently. Following the feature extraction, the processed features are fed into a four-layer vanilla transformer. This transformer is designed to perform masked prediction across the available modalities. Specifically, during the training, one of the modalities is masked, and the model is tasked with predicting the missing modality using the remaining two. When one modality is unavailable, it is filled with zeros and remains constant, simulating real-world scenarios with incomplete data.

To train the model, we use a Connectionist Temporal Classification (CTC) loss mechanism \cite{graves2012connectionist} that establishes positive and negative sample pairs. The concatenated version of the predicted and real samples from all three modalities forms a positive pair, while negative pairs are constructed by pairing each modality with data segments from different time instances. This setup allows the model to learn effective representations of physiological data and prepares the pre-trained model for downstream tasks, such as prosocial behavior intention prediction.

\subsection{Predicting Prosocial Behavior Intentions}

Figure \ref{fig:block1} outlines the process of predicting prosocial behavior intentions using the feature representations extracted from the pre-trained physiological model (described in the physiological model pre-training part of Figure \ref{fig:block1}) and body movement features. The first block in this process is the feature representation block, which takes the output embedding from the pre-trained physiological model and concatenates it with body movement feature vectors. These body movement features include head position, head rotation, and movements of the shoulders, elbows, wrists, and hands, as described earlier. To ensure that the body movement features have the same dimensions as the pre-trained physiological embedding, they are passed through a 1D CNN layer, which normalizes and matches their dimensions. The body motion signals are not included in the pre-training because it is not present in the used public multimodal datasets.

Once the feature representation block generates the concatenated feature vector, it is passed to a two-layer LSTM model. The first LSTM layer is a sequence-to-sequence model, which processes the sequential nature of the input data (captured over 5-second intervals) and outputs an intermediate sequence. The second LSTM layer operates in a sequence-to-single format, reducing the sequential output to a single prediction for each instance. Despite the latter's success in other domains, the reason for choosing LSTM over a Transformer layer is that prior literature has shown that LSTMs perform better in tasks with short-term dependencies \cite{zeyer2019comparison}, particularly in limited data settings, such as predicting prosocial behavior intentions. Even our experiments resulted in a similar observation.

Finally, the second LSTM layer output is fed into a sigmoid activation layer, which produces a binary classification (0 or 1) representing the likelihood of prosocial behavior intention, by optimizing weighted-binary cross-entropy loss (BCE). This setup enables the model to make accurate predictions based on both physiological and body movement features.

\section{Experimental Setting}
\label{ssec:Ex}

The prosocial behavior intention prediction experiment was conducted using the dataset described in Section II. Each participant completed an hour-long session, during which they encountered opportunities for prosocial behavior in driving scenarios. For each participant, we considered a 5-second window just before a prosocial behavior encounter, labeling these windows as 1 (positive interaction) if the participant performed a prosocial action. All other window frames were labeled as 0 (negative interaction). The task for the prosocial behavior intention prediction experiment was to classify a given 5-second time frame into either 1 or 0, which determines whether the participant is likely to perform a prosocial action just after the window. The features used for this task included physiological signals (HR, GSR, and pupil diameter) and body movement data (including head, shoulder, elbow, wrist, and hand positions and rotations) collected throughout the session.

To assess the impact of the proposed pre-trained physiological model on prosocial behavior intention prediction (PBIP), we compared three different experimental setups, all of which used both physiological and body movement information (referred to as All Data). In the first case (SSL-PBIP), we utilized the proposed self-supervised learning pipeline, where the embeddings from the pre-trained physiological model were concatenated with the body movement features. This combined feature set was passed to the LSTM block, as described in Section III. In the second case (LSTM-PBIP), we bypassed the pre-trained model and directly combined the raw physiological features (after processing through 1D CNN) with the body movement features before feeding them into the same LSTM architecture. To further explore alternative approaches, we considered a third case (Trans-PBIP), where we replaced the LSTM with two transformer encoder layers to examine how transformers perform for this task. Additionally, to explore the potential of physiological data alone, we introduced a special case (referred to as Only Physiological Data). In this case, we followed the same pipeline as the SSL-PBIP and LSTM-PBIP setups but provided zero values as input for the body movement data. This allowed us to isolate the impact of the physiological signals on prosocial behavior intention prediction.

For the pre-trained model, we implemented a 4-layer, encoder-only transformer, where each transformer layer had an embedding size of 128 dimensions for both the physiological features and body movement features (after passing through the 1D CNN). The model was trained on a Nvidia RTX A6000 GPU using the Adam optimizer with a learning rate of $10^{-5}$. To evaluate the performance of the models, we used two key metrics: weighted accuracy, which accounts for class imbalance, and the F1 score, which balances precision and recall to measure the model’s classification performance.

\vspace{-0.2cm}
\section{Discussion}

\begin{table}[ht]
    \centering
    \caption{Comparison of Weighted Accuracy (WA) and F1 Score for Prosocial Behavior Intention Prediction models using both All Data and Only Physiological Data. Results highlight the performance of the baseline (LSTM-PBIP and Trans-PBIP) and the proposed self-supervised learning model (SSL-PBIP).}
    \setlength{\tabcolsep}{4pt} 
    \renewcommand{\arraystretch}{1.2} 
    \begin{tabular}{ccccc}
        \hline
        & \multicolumn{2}{c}{All Data} & \multicolumn{2}{c}{Only Physiological Data} \\
        \hline
        Metrics & WA & F1 & WA & F1 \\
        \hline
        LSTM-PBIP & 0.754$\pm$0.010 & 0.729$\pm$0.011 & 0.724$\pm$0.026 & 0.694$\pm$0.013 \\
        Trans-PBIP & 0.743$\pm$0.021 & 0.728$\pm$0.013 & 0.718$\pm$0.018 & 0.683$\pm$0.006 \\
        SSL-PBIP & \textbf{0.792$\pm$0.008} & \textbf{0.753$\pm$0.013} & \textbf{0.762$\pm$0.011} & \textbf{0.741$\pm$0.008} \\
        \hline
    \end{tabular}
    \label{tab:all_results}
    \vspace{-0.5cm}
\end{table}

 To determine if the results are statistically significant, we employed a one-tailed t-test, considering significance at a $p$-value less than 0.05. This statistical analysis was crucial in confirming that the improvements across different cases were not due to random chance. From Table~\ref{tab:all_results}, it is evident that the proposed self-supervised learning model (SSL-PBIP) shows notable improvements over the baseline methods (LSTM-PBIP and Trans-PBIP) in both Weighted Accuracy (WA) and F1 score for the All Data case. All differences between the SSL-PBIP and the baseline methods were statistically significant, in both WA and F1 score. Specifically, the SSL-PBIP model achieves approximately $5\%$ higher WA and over $3\%$ improvement in F1 score compared to the LSTM-PBIP, while outperforming the Trans-PBIP model by about $6.5\%$ in WA and around $3.5\%$ in F1 score. These improvements highlight the effectiveness of leveraging a pre-trained model based on self-supervised learning (SSL) to incorporate multimodal physiological data for predicting prosocial behavior intention.

In particular, the relative performance gains for both WA and F1 metrics in the SSL-PBIP model demonstrate that pre-training with modality masking allows for better generalization when combined with body movement features. The performance difference between LSTM-PBIP and Trans-PBIP indicates that while both LSTMs and Transformers are viable architectures, the SSL-PBIP shows more than $6\%$ better WA compared to the transformer-based approach, proving the superiority of pre-training. The relative improvement in the F1 score for SSL-PBIP further emphasizes that the proposed approach not only achieves higher accuracy but also maintains a well-balanced performance across both classes, avoiding overfitting to either class. In the Only Physiological Data case, the proposed SSL-PBIP model once again achieves around $5\%$ higher WA and about $7\%$ better F1 score compared to LSTM-PBIP and Trans-PBIP baselines. It is particularly noteworthy that the proposed model with only physiological inputs performs better than the baseline models using all data, showing around $4.5\%$ improvement in WA and $6\%$ improvement in F1 score compared to the LSTM-PBIP with all data. The relatively high F1 score to that of the WA metric shows that the SSL-PBIP model is well-balanced between both classes and avoids overfitting, showing a robust solution for prosocial intention prediction with only physiological data.

\begin{figure}[t]
    \centering
    \begin{subfigure}[t]{0.7\columnwidth} 
        \centering
        \includegraphics[width=\columnwidth]{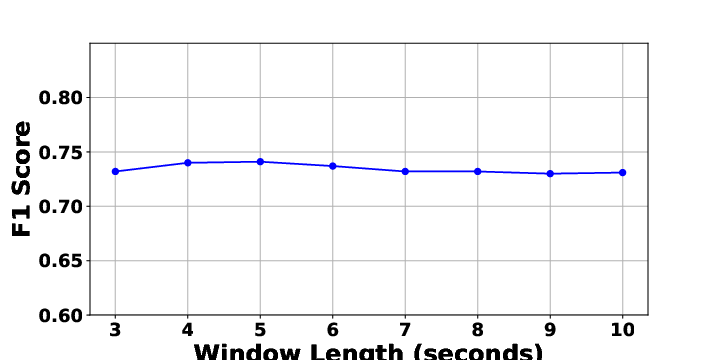}
        \caption{\footnotesize F1 Score vs Window Length.}
        \label{fig:window_length}
    \end{subfigure}
    
    \vspace{-0.1cm} 
    \begin{subfigure}[t]{0.7\columnwidth} 
        \centering
        \includegraphics[width=\columnwidth]{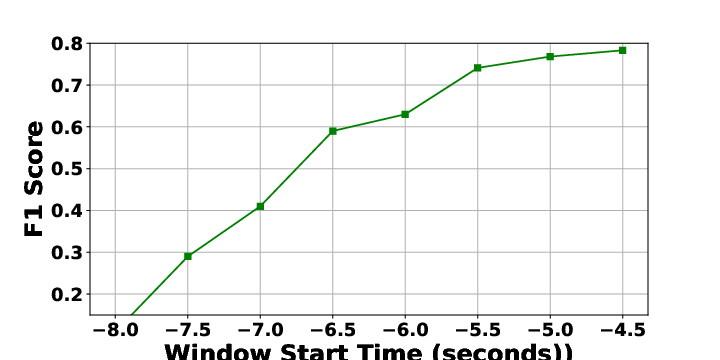}
        \caption{\footnotesize F1 Score vs Time Shift (Start Time).}
        \label{fig:start_time}
    \end{subfigure}
    
    \caption{(a) F1 Score for varying window lengths. (b) F1 Score for time shifts relative to prosocial behavior onset.}
    \vspace{-0.5cm} 
\end{figure}

The two plots in Figure \ref{fig:window_length} and Figure \ref{fig:start_time} highlight how window length and start time affect the F1 score for prosocial behavior intention prediction. In the first plot, the best performance is achieved with a 5-second window, providing a balance between capturing sufficient temporal context and avoiding irrelevant data. Shorter windows (3-4 seconds) miss key information, while longer ones dilute the significance of recent data, slightly reducing performance. The second plot examines the impact of when the window starts, with $0$ seconds marking the beginning of prosocial behavior. The results show that the most recent $2$ seconds before the behavior (starting around $t = -2$) are crucial for accurate detection. Windows starting before $t = -6$ seconds performs worse, indicating earlier data is less relevant. Notably, windows extending to $t = -4$ seconds, which capture up to $1$ second into the activity (window length is 5 seconds), still perform well due to the typical physiological delay of $0.5$ to $1$ second. This confirms that selecting the right window length and timing is essential for accurately predicting prosocial behavior intentions without capturing signals during the action itself.

\vspace{-0.2cm}
\section{Conclusions}
This paper introduces a novel self-supervised learning (SSL) approach to predict prosocial behavior intentions in mobility scenarios by pre-training on multimodal physiological data. Our proposed SSL model, which focuses on physiological signals such as heart rate, shimmer, and pupil diameter, can be applied to a wide range of tasks involving physiological signals, extending beyond prosocial behavior prediction. By leveraging existing multimodal datasets and using a transformer-based architecture to predict masked modalities, the pre-trained model offers a robust solution to the challenge of limited labeled data in this domain. We demonstrated that incorporating this pre-trained model into our prediction pipeline significantly improved performance when compared to similar pipelines without pre-training. This result highlights the potential of SSL to enhance model accuracy in tasks with small datasets, particularly in human-machine interaction and intelligent vehicle systems. Future work will involve expanding the dataset and exploring strategies to further differentiate prosocial behavior intention from explicitly avoiding prosocial situations. This will provide deeper insights into how such behaviors manifest and enhance the applicability of our approach in diverse mobility scenarios. The promising performance of the prosocial behavior intention prediction paves the road for real-life automated vehicle and service robot applications, to improve the safety, efficiency, and user experiences of mobility.

\bibliographystyle{IEEEtran}
\bibliography{mybib}

\end{document}